\documentclass[conference]{IEEEtran}

\usepackage{filecontents,amsmath,graphicx}

\IEEEoverridecommandlockouts

\title{A Study on Effects of Implicit and Explicit Language Model Information for DBLSTM-CTC Based Handwriting Recognition}

\author{\IEEEauthorblockN{Qi Liu\IEEEauthorrefmark{1}\IEEEauthorrefmark{2}\thanks{*This work was done when Qi Liu was an intern in Speech Group of Microsoft Research Asia, Beijing, P. R. China.},
Lijuan Wang\IEEEauthorrefmark{2},
Qiang Huo\IEEEauthorrefmark{2} }
\IEEEauthorblockA{\IEEEauthorrefmark{1}ACM Honored Class, Zhiyuan College, Shanghai Jiao Tong University,
Shanghai 200240, P. R. China}
\IEEEauthorblockA{\IEEEauthorrefmark{2}Microsoft Research Asia, Beijing 100080, P. R. China \\
Emails: liuq901@gmail.com; \{lijuanw, qianghuo\}@microsoft.com}
}

\begin{document}

\maketitle

\begin{abstract}
Deep Bidirectional Long Short-Term Memory (DBLSTM) with a Connectionist Temporal Classification (CTC) output layer has been established as one of the state-of-the-art solutions for handwriting recognition. It is well-known that the DBLSTM trained by using a CTC objective function will learn both local character image dependency for character modeling and long-range contextual dependency for implicit language modeling. In this paper, we study the effects of implicit and explicit language model information for DBLSTM-CTC based handwriting recognition by comparing the performance of using or without using an explicit language model in decoding. It is observed that even using one million lines of training sentences to train the DBLSTM, using an explicit language model is still helpful. To deal with such a large-scale training problem, a GPU-based training tool has been developed for CTC training of DBLSTM by using a mini-batch based epochwise Back Propagation Through Time (BPTT) algorithm.
\end{abstract}

\section{Introduction}
Long Short-Term Memory (LSTM) \cite{hochreiter1997long,gers2000learning,gers2003learning} is a special type of Recurrent Neural Networks (RNN) (e.g., \cite{williams1995gradient}), which has been used to build handwriting recognition (HWR) systems for a long time by using a single-hidden-layer Bidirectional LSTM (BLSTM) \cite{graves2009-PAMI} or a deep Multi-Dimensional LSTM (MDLSTM) \cite{graves2009-MDLSTM}, both with a so-called Connectionist Temporal Classification (CTC) output layer and trained with a CTC-based objective function \cite{graves2006-CTC}. Recently, BLSTM-CTC approach was also applied successfully to optical character recognition (OCR) for printed text (e.g., \cite{ocr-BLSTM-icdar2013}), while MDLSTM-CTC approach was used to build several state-of-the-art offline HWR systems (e.g., \cite{bluche2014a2ia,pham2014dropout,Moysset-ICFHR2014}). Furthermore, BLSTM was used as a feature extractor to build Gaussian mixture hidden Markov model (GMM-HMM) based offline HWR systems, which achieved state-of-the-art performance on several benchmark tasks (e.g., \cite{Kozielski-ICDAR2013,Kozielski-DAS2014,Hamdani-DAS2014}). Inspired by the success of using Deep BLSTM (DBLSTM) and HMM (DBLSTM-HMM) for automatic speech recognition (ASR) \cite{graves2013hybrid}, more recently, DBLSTM-HMM approach has been used to build HWR systems with promising results (e.g., \cite{Doetsch-ICFHR2014,Voigtlaender-ICASSP2015,Zhang-ICDAR2015,Chen-ICDAR2015}). Finally, DBLSTM-CTC approach has been tried with state-of-the-art performance on Rimes and IAM benchmark tasks \cite{Bluche-DBLSTM-CTC-2014}, which is also the approach studied in this paper.

It is well-known that the (D)BLSTM and MDLSTM trained by using a CTC objective function will learn both local character image dependency for character modeling and long-range contextual dependency for implicit language modeling. In \cite{ocr-BLSTM-icdar2013}, it is demonstrated that very high OCR accuracy can be achieved for printed text by a BLSTM-CTC recognizer trained with about 95k text line images but without using any explicit language model (LM) in decoding. However, for more difficult handwriting recognition tasks, an explicit LM is typically used to achieve higher character/word recognition accuracies in DBLSTM-CTC or MDLSTM-CTC based HWR systems (e.g., \cite{bluche2014a2ia,Moysset-ICFHR2014,Bluche-DBLSTM-CTC-2014}). So far, the scale of training set for HWR experiments reported in literature is relatively small, ranging from several to tens of thousands of text lines. It is not clear yet what would happen when a much larger training set could be used. In this paper, we study the effects of implicit and explicit language model information for DBLSTM-CTC based handwriting recognition. First, we train several sets of DBLSTM using different amount (up to one million text lines) of training data. Then, we conduct recognition experiments for each DBLSTM-CTC based recognizer by using and not using an explicit LM in decoding. The LMs include character $n$-gram and word trigram.
By comparing and analyzing the experimental results, we hope to gain some insights. To deal with such a large-scale training problem, we have developed a GPU-based training tool for CTC training of DBLSTM by using a mini-batch based epochwise Back Propagation Through Time (BPTT) algorithm (e.g., \cite{williams1995gradient}).

The rest of this paper is organized as follows: Section II introduces how we build a DBLSTM-CTC based HWR system. Section III presents experimental results. Finally, we conclude the paper in Section IV.

\section{Overview of DBLSTM-CTC based HWR System}

\subsection{Preprocessing and Feature Extraction}
Given each horizontal text line image, preprocessing steps of baseline and slant correction will be applied first. An approach similar to that in \cite{website:rlsaprojection} is used, which is based on Run Length Smoothing Algorithm (RLSA) (\cite{wahl1982block,wahl1983new}) and projection profile based techniques (e.g., \cite{vinciarelli2001new,pastor2004projection}). Images are rotated by angles within a certain range and then smoothed by RLSA. Each rotation is evaluated by different objective functions to find an optimal angle for baseline and slant correction. After baseline and slant correction, each horizontal text line image is normalized to have a height of 60 pixels.

For feature extraction, each sentence is first split into frames by a sliding window of 30 pixels wide with a frame shift of 3 pixels. Then each frame is smoothed by applying a horizontal cosine window to derive a 1,800-dimensional raw feature vector. The dimension of raw feature vectors is reduced to 50 by principal component analysis (PCA). Finally, these 50-dimensional feature vectors are normalized such that each dimension of feature has a zero sample mean and unit sample variance on training set.

\subsection{DBLSTM-based Character Modeling and CTC Training}
Given the set of training feature vector sequences with transcriptions, a DBLSTM with a CTC output layer can be trained by using a CTC-based objective function as described in \cite{graves2009-PAMI,graves2006-CTC}. The CTC output layer is a softmax layer with 79 classes, including 52 case-sensitive English letters, 10 digits, 15 punctuation marks, a ``space", and a special ``blank'' symbol. ``Blank'' is not a real character class, but a virtual symbol used to separate the consecutive real symbols. For a given feature vector sequence, the output of the DBLSTM at each time-step gives the corresponding posterior probability distribution of 79 classes. The memory block of DBLSTMs has the same topology as that in \cite{graves2009-PAMI,graves2013hybrid}.

For a small-scale training set, it is feasible to use the RNNLIB open source toolkit \cite{RNNLIB} for CTC training of DBLSTM, which is a single-thread CPU-implementation of an epochwise BPTT algorithm. However, for the large-scale training set of one million text lines we are dealing with, it would take several months to train a DBLSTM by using the RNNLIB tool, therefore be infeasible to conduct any meaningful experiments. To support training with big data, we have developed a GPU-based training tool for CTC training of DBLSTM by using a mini-batch based epochwise BPTT algorithm, which does not need frame-level ground-truth labels. It is noted that the open-source tool of CURRENNT \cite{CURRENNT} has also implemented a mini-batch based epochwise BPTT algorithm, but only supports frame-level training which requires the target label for each frame. In our case, we adopt a CPU-GPU cooperative implementation. We let CTC-related code run on CPU and other code run on GPU. It is because the CTC code is hard to parallelize and incurs much memory overhead, yet the CTC part is not the bottleneck of the whole pipeline. We also transfer the whole array between the CTC code and the other code to reduce CPU-GPU memory communication. During the development process, we have also implemented a single-thread CPU version of CTC training tool for DBLSTM, which is much more efficient than RNNLIB, yet can achieve similar recognition accuracy. Our single-GPU implementation can achieve about 30 times speedup in comparison with our CPU implementation.

\subsection{Language Model}
We have used SRILM toolkit \cite{stolcke2002srilm} and a text corpus from Linguistic Data Consortium (LDC) (catalog number LDC2008T15) \cite{ldccorpus} to build several language models, i.e., character $n$-gram ($n=3,4,5,8,10$) and word trigram, with Good-Turing discounting. 500 (out of 4495) documents of the LDC corpus are used. For character $n$-gram,
the vocabulary consists of 78 real character classes (without special ``blank'' symbol). For word trigram, the vocabulary consists of 200k words with top occurring frequencies in the training corpus, which leads to an out-of-vocabulary (OOV) rate of 8\% on IAM-online test set \cite{liwicki2005-IAM-online,website:online-iam-partition}.

\begin{table*}[t]
\caption{Statistics of several training sets derived from a Microsoft handwriting corpus.}
\label{statistics-Microsoft-data}
\centering
\begin{tabular}{|c|c|c|c|c|c|c|c|c|c|}
\hline
\bfseries \# of Text Line Images & 10K & 20K & 50K & 80K & 100K & 200K & 500K & 800K & 1M \\
\hline
\bfseries \# of Words & 69,144 & 138,344 & 346,139 & 543,567 & 673,421 & 1,192,101 & 2,754,690 & 4,231,015 & 5,268,676  \\
\hline
\bfseries \# of Characters & 352,156 & 704,985 & 1,761,416 & 2,759,539 & 3,462,030 & 6,153,498 & 16,894,999 & 27,452,759 & 32,717,204  \\
\hline
\bfseries \# of Unique Text Lines & 5544 & 10677 & 12611 & 12930 & 15686 & 18654 & 41685 & 43997 & 44053 \\
\hline
\end{tabular}
\end{table*}

\begin{table*}[t]
\caption{Training time of DBLSTMs using GPU-based tool on different sets of Microsoft training data.}
\label{large}
\centering
\begin{tabular}{|c|c|c|c|c|c|c|c|c|c|}
\hline
\bfseries Dataset & 10K & 20K & 50K & 80K & 100K & 200K & 500K & 800K & 1M \\
\hline
\bfseries Average Time per Epoch (sec.) & 299 & 526 & 1,251 & 1,933 & 2,430 & 4,273 & 10,983 & 17,559 & 21,010 \\
\hline
\bfseries Number of Epoches & 33 & 25 & 36 & 29 & 29 & 35 & 45 & 42 & 42 \\
\hline
\bfseries Total Time (day) & 0.11 & 0.15 & 0.52 & 0.64 & 0.81 & 1.73 & 5.72 & 8.53 & 10.21 \\
\hline
\end{tabular}
\end{table*}

\subsection{Decoding}

We have tried the following two decoding methods:

\subsubsection{CTC-based Decoding without using LM}
This method is called best path decoding in \cite{graves2006-CTC} and works as follows. Given the feature vector sequence of an unknown text line, as mentioned above, the CTC output at each time-step gives a probability distribution over a set of symbols. At each time-step, choose the symbol with the highest probability. Then, concatenate the most active outputs at every time-step to obtain a raw output sequence $S$. Finally, apply a merge function, $\beta(S)$, to generate the final decoding result. The merge function $\beta(S)$ maps the symbol sequence $S$ to another label sequence by first combining the same consecutive symbols together and then removing all the ``blank'' symbols. For example, $\beta(a--aab)=\beta(--a-ab)=\beta(a-abbb)=\beta(aa-aab)=aab$, where $-$ means ``blank''.

Best path decoding is easy to implement. It only leverages the implicit language model information embedded in DBLSTM. Because no word-lexicon is used, there is no OOV issue. From our HWR experimental results to be reported later, this decoding method can achieve a reasonable character error rate (CER), but a relatively high word error rate (WER). There are much more words containing few character errors in each word than the words containing many character errors in each word.

\subsubsection{WFST-based Decoding with LM}
This is the method used in, e.g., \cite{bluche2014a2ia,pham2014dropout,Moysset-ICFHR2014,Bluche-DBLSTM-CTC-2014}. The decoder searches a graph based on Weighted Finite-State Transducers (WFST) constructed from several main system components. Kaldi toolkit \cite{Kaldi} is used for both WFST construction and decoding.

For a given feature vector sequence $X$ of an unknown text line, at each time-step, the CTC-trained DBLSTM provides a probability distribution $P(s|X)$ over the symbol set. If we model each symbol by a single-sate HMM with a self-loop and outgoing transitions, its ``state-dependent" likelihood score can be approximated by $\frac{P(s|X)}{P(s)^\alpha}$, where $P(s)$ is the prior distribution of symbol $s$ estimated from the transcriptions of the training text line images with a special treatment of ``blank" symbol, and $\alpha$ is a tunable scaling parameter ($\alpha=0.2$ in our experiments). These HMMs are transformed into a WFST $H$. If a word lexicon is used, we can decompose each word by inserting an optional ``blank" symbol between two different characters and a compulsory ``blank" symbol between two repeated characters. Then a lexicon FST $L$ can be constructed in the structure of a WFST. A character $n$-gram or a word trigram LM can also be transformed into a WFST $G$. Depending on which LM to use, the final search graph can be constructed by composing a WFST of $HLG$ or $HG$.

Once the search graph is built, the decoder will take the sequence of ``state-dependent" likelihood scores as input and generate a word sequence as recognition result.

%\begin{table}[t]
%\caption{The Speed of the System on Small Scale Data}
%\label{speed}
%\centering
%\begin{tabular}{|c|c|c|c|}
%\hline
%\bfseries Dataset & \bfseries Small IAM & \bfseries Standard IAM & \bfseries Enlarged IAM\\
%\hline
%\bfseries CPU Speed (sec) & 1037 & 6033 & 53241\\
%\hline
%\bfseries GPU Speed (sec) & 41 & 200 & 1467\\
%\hline
%\bfseries Speedup & 25.2x & 30.1x & 36.2x\\
%\hline
%\end{tabular}
%\end{table}

\begin{table}[t]
\caption{Effects of learning rate (LR) scheduling on IAM-offline dataset. CTC-based decoding without LM is used.}
\label{ler}
\centering
\begin{tabular}{|c|c|c|c|c|}
\hline
\bfseries \# of Times for LR Reduction & 0 & 3 & 6 & 9 \\
\hline
\bfseries Character Error Rate (\%)& 20.8 & 16.5 & 15.7 & 15.7 \\
\hline
\end{tabular}
\end{table}

\begin{table*}[!t]
\caption{Effects of learning rate (LR) scheduling on Microsoft datasets. CTC-based decoding without LM is used (CER/WER in \%).}
\label{first}
\centering
\begin{tabular}{|c|c|c|c|c|c|c|c|c|c|}
\hline
\bfseries Dataset & 10K & 20K & 50K & 80K & 100K & 200K & 500K & 800K & 1M \\
\hline
\bfseries Without LR Reduction & 42.8/84.8 & 30.1/72.8 & 22.5/61.9 & 19.2/56.4 & 18.7/55.1 & 17.5/53.4 & 16.2/50.3 & 16.1/50.9 & 16.2/54.2 \\
\hline
\bfseries With 3 Times LR Reduction & 42.8/85.7 & 28.0/69.9 & 21.0/59.3 & 17.9/54.2 & 16.7/51.0 & 15.0/47.5 & 14.1/45.1 & 13.5/44.0 & 13.2/44.6 \\
\hline
\bfseries With 6 Times LR Reduction & 44.0/89.6 & 27.5/68.9 & 20.0/56.8 & 17.0/51.1 & 15.8/48.2 & 14.0/44.1 & 12.9/42.1 & 12.3/40.4 & 11.8/40.1 \\
\hline
\end{tabular}
\end{table*}

\begin{table*}[!t]
\caption{Performance (CER/WER in \%) comparison of CTC-based decoding without using LM and WFST-based decoding with different types of LMs.}
\label{second}
\centering
\begin{tabular}{|c|c|c|c|c|c|c|c|c|c|}
\hline
\bfseries Dataset & 10K & 20K & 50K & 80K & 100K & 200K & 500K & 800K & 1M \\
\hline
\bfseries CTC Decoding w/o LM & 44.0/89.6 & 27.5/68.9 & 20.0/56.8 & 17.0/51.1 & 15.8/48.2 & 14.0/44.1 & 12.9/42.1 & 12.3/40.4 & 11.8/40.1 \\
\hline
\bfseries Character 3-gram  & 42.7/89.8 & 26.6/81.3 & 17.9/75.2 & 15.2/72.9 & 14.3/72.1 & 12.5/69.9 & 11.3/68.2 & 10.9/67.2 & 10.1/67.1 \\
\hline
\bfseries Character 4-gram  & 41.0/80.5 & 24.3/65.1 & 15.8/55.8 & 13.4/52.1 & 12.6/50.7 & 11.1/48.4 & 9.9/45.7 & 9.7/44.5 & 8.8/44.5 \\
\hline
\bfseries Character 5-gram  & 40.0/76.1 & 22.3/56.9 & 14.1/47.3 & 11.7/43.0 & 11.1/41.7 & 9.5/39.5 & 8.5/36.6 & 8.5/34.9 & 7.7/35.2 \\
\hline
\bfseries Character 8-gram  & 39.8/75.3 & 21.7/55.1 & 13.1/44.7 & 10.9/41.1 & 10.3/39.4 & 8.9/37.9 & 7.8/34.7 & 8.0/32.5 & 7.0/33.5 \\
\hline
\bfseries Character 10-gram  & 40.3/75.6 & 22.5/55.5 & 13.3/44.7 & 11.2/41.2 & 10.6/39.4 & 9.1/37.9 & 8.1/34.5 & 8.1/32.0 & 7.3/33.6 \\
\hline
\bfseries Word Trigram & 34.8/59.0 & 17.0/34.6 & 11.1/27.5 & 9.2/24.1 & 8.6/22.9 & 7.7/21.2 & 6.8/19.6 & 6.2/18.1 & 6.0/18.0 \\
\hline
%\bfseries Word Trigram (RWTH) & 36.5/61.0 & 12.7/27.2 & 10.1/24.8 & 8.1/21.2 & 7.6/20.0 & 6.8/18.8 & 6.1/17.0 & 5.5/15.6 & 5.3/15.7 \\
%\hline
\end{tabular}
\end{table*}

\section{Experiments}

\subsection{Experimental Setup}
Several handwriting corpora are used in our experiments. The first one is IAM offline handwritten English sentence dataset, which contains 6,159 training and 1,861 testing text line images, respectively \cite{marti2002-IAM-offline,website:offline-iam-partition}. We use this small dataset primarily for developing and debugging our CPU- and GPU-based training tools so that RNNLIB tool can be used for comparison purpose.

The second corpus is a large-scale Microsoft in-house ink (i.e., online handwritings) database, which contains more than one million online handwritten English text lines. We render each line of online handwriting into a text line image. More specifically, we model each stroke by a B{\'e}zier curve. For every four consecutive points, we connect them by a cubic B{\'e}zier curve. The thickness of B{\'e}zier curve controls the thickness (3 pixels here) of the stroke. From such a corpus of rendered handwritten text line images, 9 sets of training data with different sizes are formed and some statistics are summarized in Table \ref{statistics-Microsoft-data}.

To test HWR systems built from Microsoft datasets, we use the test set of IAM online handwritten English sentence dataset, which contains 3,859 text lines with 20,272 words and 89,153 characters in total \cite{liwicki2005-IAM-online,website:online-iam-partition}. We use the same method to render testing text line images as in rendering training data from Microsoft ink data.

Learning rate (LR) scheduling is important for neural network training. Lower learning rate often gives better result but higher learning rate needs less epoches to converge. Therefore, we use the following simple scheme for learning rate scheduling: First, train the network with a high learning rate such as $10^{-3}$ or $10^{-4}$ for several epoches. Then, cut the learning rate in half, retrain the network. At this time, only one or two epoches is enough. Finally, repeat this cut-down and retrain procedure until the learning rate reaches a low value such as $10^{-5}$ or $10^{-6}$. For gradient-based BPTT training, the momentum is set as 0.9 and the initial learning rate is $10^{-4}$.

For large-scale experiments on Microsoft datasets, we use a fixed configuration. The DBLSTM has 5 hidden BLSTM layers, each has 240 memory cells (120 for forward and 120 for backward states). All experiments are run on a server with Intel Xeon CPUs, 128GB memory, and an NVIDIA Tesla K20Xm GPU. The Operating System of the server is Microsoft Windows Server 2012. We use both CER and WER as performance metrics.

\subsection{Experimental Results}

To compare the efficiency of our CPU- and GPU-based training tools, we measure the wall-clock time for running a single epoch of training DBLSTM on the IAM-offline training set. It takes 6,033 and 200 seconds for CPU- and GPU-based tools respectively. GPU-based tool achieves about 30 times speedup. Both tools can train DBLSTMs leading to similar recognition accuracy, therefore we use our GPU-based training tool for the remaining experiments. Table \ref{large} shows training time of DBLSTMs using our GPU-based tool on different sets of Microsoft training data. The total time scales almost linearly with the amount of training data. For the largest dataset with one million lines, it takes about 10 days to complete training using a single GPU card.

 Table \ref{ler} shows the CER of the CTC-based decoding without using LM on the IAM-offline testing set with different times of LR reduction. Table \ref{first} shows the CER/WER (in \%) of the CTC-based decoding without using LM on the IAM-online testing set achieved by DBLSTMs trained from different sets of Microsoft data with different LR scheduling. The experimental results indicate that the LR scheduling is helpful for both cases, and is more important for larger scale training tasks.
\begin{figure}[!t]
\centering
\includegraphics[width=3.5in]{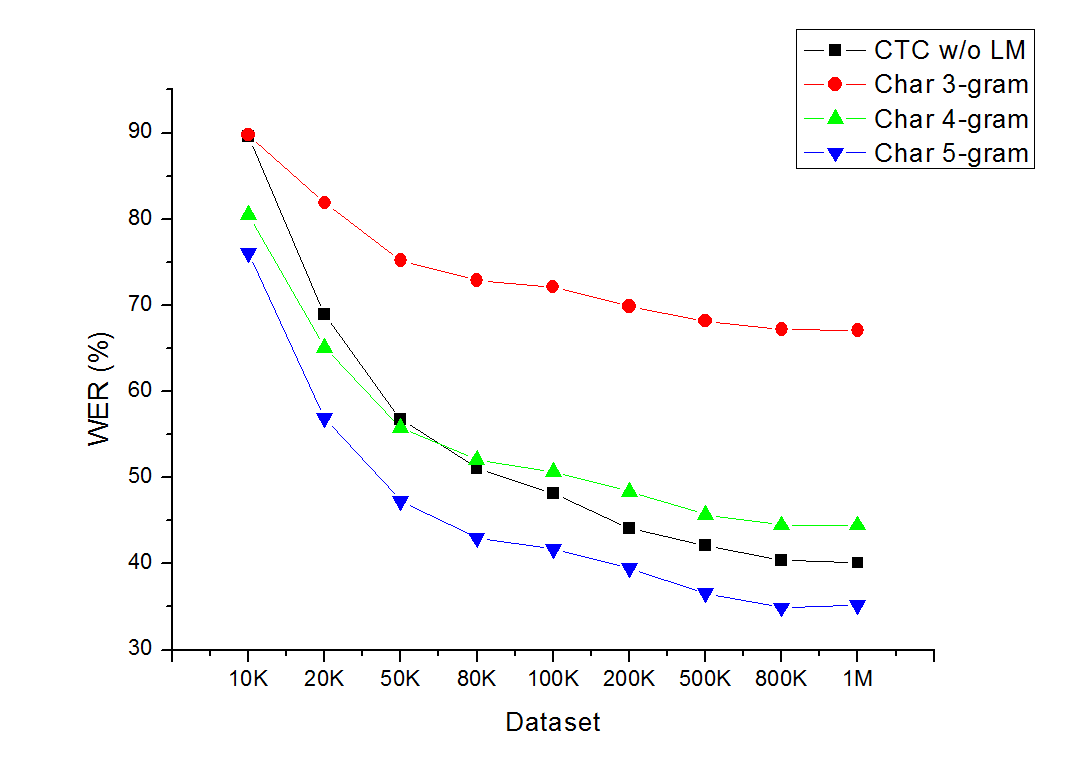}
\caption{Effects of Implicit LM vs. Character $n$-gram.}
\label{fig}
\end{figure}

Table \ref{second} summarizes the performance (CER/WER in \%) comparison of CTC-based decoding without using LM and WFST-based decoding with different types of LMs. Figure \ref{fig} compares the WERs of CTC-based decoding without using LM (i.e., implicit LM information is used) and WFST-based decoding with different character $n$-gram LMs. Several observations can be made. First, for CTC-based decoding without using LM, both CER and WER improves with the increasing amount of training data. The improvement comes from both the improved modeling for local character image modeling and the improved implicit language modeling. From Figure \ref{fig}, it is clear that CTC-based decoding without using LM performs as well as WFST-based decoding with a character 3-gram LM initially, then performs similarly with the WFST-based decoding with a character 4-gram LM when more training data is used, and then outperforms that of using character 4-gram LM when even more training data is used, but cannot surpass the WFST-based decoding with a character 5-gram LM even when about 1M lines of training data is used. Second, for CTC-based decoding without using LM, a CER of 11.8\% can be achieved for 1M training case, but the corresponding WER is only 40.1\%. The CER and WER can be reduced to 7.0\% and 33.5\% respectively by using WFST-based decoding with a character 8-gram LM. This shows clearly the effectiveness of using a powerful explicit LM. However, using character 10-gram does not bring additional improvement. Third, the CER and WER can be further reduced to 6.0\% and 18.0\% respectively by using WFST-based decoding with a word trigram. The relatively small improvement of CER (14\% relative CER reduction) and much bigger improvement of WER (46\% relative WER reduction) shows clearly the power of lexical constraints.

\section{Conclusion and Discussions}
From the above results, we conclude that even using one million lines of training sentences to train the DBLSTM, using an explicit word trigram language model is still very helpful. Actually, if we use a better word trigram LM shared by RWTH team \cite{Kozielski-ICASSP2013} in WFST-based decoding, the CER and WER can be further reduced to 5.3\% and 15.7\% respectively. 

It is not surprised at all that the result of CTC-based decoding without using LM outperforms the WFST-based decoding using a weak character $n$-gram, simply because CTC-trained DBLSTM has learned an implicit LM from the training data already. How powerful such an implicit LM would really depends on the amount and nature of the training data. A careful analysis of our training data reveals that there are only about 44k unique text lines out of about 1M lines of training text lines, therefore the learned implicit LM is not general and strong enough. That explains partially why a strong character $n$-gram LM or an even stronger word trigram LM could help a lot.

Our ongoing and future works include 1) improving preprocessing techniques for baseline and slant correction, 2) improving language model, 3) developing new discriminative training method for sequence training of DBLSTM, 4) using even more representative training data to improve models.

\section*{Acknowledgement}
The authors would like to thank Professor Hermann Ney and his team at RWTH Aachen University to share their language model for our experiments.

\end{document}